# BEAMS: Benchmarking and Evaluating AI for Modeling and Simulation

By: Sara Metcalf and William Schoenberg

smetcalf@buffalo.edu
bschoenberg@iseesystems.com

## Abstract

AI tools to support real-world decision making must be able to build simulation models that inform their recommendations and render them interpretable. Tools that can automate aspects of modeling practice must complement human expertise, not replace it. The BEAMS Initiative aims to guide the development of AI tools for modeling and simulation toward forms that are responsible and ethical by establishing benchmarks for human-centered modeling and simulation practices. The initiative uses open digital and organizational infrastructure to collaboratively evaluate AI tools for modeling and simulation. The open source sd-ai project hosted by the initiative establishes transparency and enables contributions to be shared broadly. A steering group focuses on prioritizing potential benchmarks, while a technical group focuses on implementing the benchmarks in the form of automated tests. Tests for several distinct categories of evaluation have been implemented and applied to AI tools that support qualitative model building, quantitative model building, and model discussion. These include tests for causal translation, model iteration, causal reasoning, conformance, model behavior explanation, suggested model building steps, and suggested model fixes. When engines from the sd-ai project are coupled with different LLMs, their performance on these evaluations reveals variability across different AI tools. The evaluations implemented by the initiative demonstrate that AI-enabled modeling tools perform better at discussion and basic qualitative tasks than with causal reasoning and quantitative error fixing. No single LLM dominates across engine types, highlighting the importance of specific tasks and tradeoffs between speed and accuracy. Ongoing efforts of the initiative aim to incorporate benchmarks that address concerns about bias by considering alternative perspectives and human-centered use cases.

## 1. Introduction

As artificial intelligence (AI) becomes more deeply intertwined with the way people make sense of complex problems, perspectives are increasingly polarized over its potential benefits and harms. While adherents of generative AI may point to transformational efficiency gains, skeptics may point to opacity of the internal processes underlying responses to user queries (Hosseini et al., 2025). We contend that for its benefits to be realized, AI must be interpretable, transparent, and understandable. AI tools whose outputs cannot be easily verified, whose reasoning is opaque, or whose conclusions cannot be trusted are inadequate for informing real-world decisions about pressing societal problems. Simulation modeling is fundamentally about sense-making, by representing the structures that generate the behaviors we observe in the real world. We maintain that AI cannot be wielded responsibly to address societal challenges unless it explicitly constructs simulation models, as it is through simulation models that we can apply established analytical methods to understand what an AI tool is doing, why it recommends a particular option, and how it arrives at a given conclusion.

The proliferation of generative AI tools in the form of large language models (LLMs) has inspired modeling and simulation researchers to explore the capability of these tools for automating aspects of the modeling process (du Plooy & Oosthuizen, 2023; Hosseinichimeh et al., 2024; Vanhee et al., 2025). Giabbanelli (2023) shows the potential value of LLMs for helping people understand and interpret conceptual and computational models with respect to explaining model structure, communicating simulation results, and providing constructive feedback to resolve model errors.

As AI-enabled tools become increasingly available to modelers and begin to appear in simulation software platforms, the need to establish generalizable guidelines for evaluating them becomes more acute. The Benchmarking and Evaluating AI for Modeling and Simulation (BEAMS) Initiative[1] was organized to address this need through an open collaboration among information scientists, software developers, modelers, and model users to establish benchmarks for how well AI-enabled tools can support the practice of modeling. By setting clear evaluation criteria for AI-enabled modeling tools, the BEAMS Initiative seeks to incentivize the development of more responsible and ethical AI tools that can advance the state of the art in simulation modeling to solve critical societal problems.

The BEAMS Initiative is motivated by the following questions:

1. How do we drive LLM builders to make responsible and ethical AI tools that suit the modeling and simulation community?
2. How do we (start) to measure the performance of AI tools for modeling and simulation tasks?

Evaluation of LLM performance in tasks of both constructing and interpreting model structures such as causal maps may be conducted using reference-based or reference-free methods, depending on whether AI-enabled model outputs are assessed relative to a known ground truth (Giabbanelli et al., 2025). Building on these evaluation approaches, it becomes equally important to consider not only how accurately AI systems generate or interpret model structures, but also how they should be positioned within the broader modeling process, specifically, as tools that enhance rather than supplant the expertise of human modelers (du Plooy & Oosthuizen, 2023; Vanhee et al., 2025). Positioning AI as a complement, not a replacement, for the human modeler means designing tools that amplify expert judgment, contextual reasoning, and ethical scrutiny throughout the modeling lifecycle. In practice, this looks like using LLMs to accelerate routine steps (e.g., drafting causal links, documenting assumptions) while keeping the modeler "in the loop" to challenge premises, reconcile stakeholder perspectives, and validate structure/behavior against purpose. Evidence underscores this balance: AI tools vary widely in technical correctness and instruction conformance across LLMs, which makes expert oversight indispensable for interpreting edge cases, resolving errors, and adjudicating trade-offs between accuracy and cost (Schoenberg et al., 2025; Schoenberg et al., 2026).

[1] https://www.buffalo.edu/ai-data-science/research/beams.html

This human-AI collaboration aligns with broader findings in the literature (Muthia et al., 2025); recent studies explore automating aspects of conceptualization and explanation (du Plooy & Oosthuizen, 2023; Giabbanelli et al., 2023; Armenia et al., 2024; Ghaffarzadegan et al., 2024; Jalali & Akhavan, 2024; Liu & Keith, 2024; Veldhuis et al., 2024; Giabbanelli et al., 2025; Hu, 2025, Schoenberg, 2026), but they also flag persistent challenges in traceability, evaluation, and stakeholder engagement, areas where human modelers' systems thinking and domain knowledge are critical. Treating AI as a cognitive aid helps the community pursue larger, more complex questions without diluting rigor, a stance that the BEAMS Initiative operationalizes by benchmarking AI tools against transparent, testable criteria to encourage responsible, ethical development for real-world decision support.

# 2. Approach

The approach taken in this research involved creation of both digital and organizational infrastructure to support collaboration to develop and share evaluations of AI-enabled tools for modeling and simulation.

## 2.1 Open Source Platform

The digital infrastructure of the sd-ai project[2] on GitHub provides an open source platform to support development of AI-enabled tools called "engines" for modeling and to support development of tests that evaluate the performance of those tools (Schoenberg et al., 2026). The sd-ai project provides reusable and extensible infrastructure in the form of a request handler that, coupled with an engine, forms an interface to facilitate communication between a modeler working within a software client and an external AI tool by providing context appropriate to the expected model output. The sd-ai platform utilizes structured outputs to support an efficient communication schema and ensure that the response from the external AI tool is in parseable JSON format.

The sd-ai project is available under the open source MIT license so that the broader modeling community can benefit from transparency about the approaches and technical details behind the different engines and evaluations.

[2] https://github.com/UB-IAD/sd-ai

## 2.2 Open Collaboration

The BEAMS Initiative was formally launched in February 2025 through the Institute for Artificial Intelligence and Data Science (IAD) at the University at Buffalo (UB). The mission of the BEAMS Initiative, which is to foster the development of responsible and ethical AI tools for modeling and simulation through open collaboration on benchmarks, is aligned with IAD's aim of bringing people together to tackle complex societal problems through advancements in AI, data science, computational science, and related areas of research.

Beginning in April 2025, BEAMS has hosted the digital infrastructure of the sd-ai project on GitHub. Anchoring the initiative in an academic institution ensures that the infrastructure is not specific to any particular software vendor, making it easier for developers of different software platforms to get involved. And though it is based at UB, the BEAMS Initiative extends beyond UB to include industry professionals as well as academics from other institutions.

The BEAMS Initiative comprises two working groups: a steering group and a technical group. The steering group focuses on the broader direction of the BEAMS initiative and prioritizing the benchmarks to be implemented, while the technical group focuses on implementing the benchmarks in the form of automated tests for the sd-ai project. The working groups have met virtually on a monthly basis since June 2025, and some BEAMS collaborators choose to participate in both groups.

At the kickoff meeting for the BEAMS Initiative in May 2025, a general process was proposed by which the steering and technical groups could develop benchmarks for AI-enabled modeling tools. This process would involve the steering group articulating *principles* for how AI-enabled automation should improve dynamic modeling, then both groups agreeing on *goals* for what aspects of AI-enabled tools and their outputs could be measured and evaluated to spur their development in a way that improves dynamic modeling, and then the technical group devising *tests* to measure those aspects. Even for evaluations already implemented, the alignment of principles, goals, and tests has provided a useful framework for BEAMS in taking stock of its evaluation suite and shaping future priorities.

The BEAMS Initiative has articulated the following design principles for how AI-enabled tools should improve dynamic modeling:

- Do no harm
- Develop tools that help people
- Increase access to the modeling process for all
- Complement human abilities, don't substitute for the human
- Provide information without bias, stereotypes, or generalizations
- Work with the modeling process, don't bypass it
- Deliver high quality models
- Use appropriate information in building models

As BEAMS collaborators consider the question of how well AI tools support the modeling process, the linkage between principles and tests increasingly centers on identifying and measuring aspects of AI performance that are most relevant to the potential role of an AI tool in enhancing modeling practice. For this exercise, potential evaluations of tasks are organized by stage of a generalized modeling process: 1) problem definition; 2) conceptualization and model architecture; 3) model formulation and implementation; 4) model analysis and testing; and 5) policy analysis and testing.

# 3. Results

## 3.1 Evaluation Categories and Tests

To date, the BEAMS Initiative has developed tests for three different types of AI-enabled modeling tools, or engines. Several categories of evaluation have been defined, each consisting of multiple distinct tests. Evaluations have been developed for engines that: 1) build qualitative models, 2) build quantitative models, and 3) discuss models. Four evaluation categories apply to AI-enabled engines that build either qualitative models (i.e., causal loop diagrams) or quantitative (i.e., stock-flow) models: causal translation, model iteration, causal reasoning, and conformance to user instructions. The quantitative engine can also be evaluated in terms of its ability to identify and fix errors in a given model. For AI-enabled engines that discuss quantitative models, there are three categories of evaluation of the tool’s ability to aid in the modeling process: model behavior explanation, suggested model building steps, and suggested model fixes. Each of these categories of evaluation is described in detail below.

### 3.1.1 Causal Translation

The causal translation evaluation (Schoenberg et al., 2026) aims to objectively assess an AI-enabled tool’s capacity to transform plain-language descriptions into structured representations of simulation models. It achieves this by employing “fake alternate universes” with fully synthetic ground truth, ensuring a single correct interpretation while reducing the likelihood of the AI-enabled tool from drawing on prior knowledge. This approach enables a controlled and rigorous assessment of a tool's most basic ability to build simulation models. It does not include any complex sentence structures or red-herrings, i.e. correlations instead of casualties. To construct the ground truth for each test, the algorithm generates both natural language causal descriptions and the corresponding models they represent. The evaluation generation algorithm starts with a list of gibberish nouns, which are uniformly pluralized to serve as variable names. Simple causal sentences are then created via a deterministic algorithm describing model structure in a simple but generally human appearing manner. There are 24 qualitative causal translation tests and nine (9) quantitative causal translation tests for AI-enabled modeling tools. Tests for causal translation align with the principle of “deliver high quality models” via the goal of accuracy.

### 3.1.2 Model Iteration

The model iteration evaluation aims to assess an AI-enabled tool's ability to extend an existing causal model by correctly adding new causal relationships without breaking the structure that is already present. Like the causal translation evaluation, this evaluation uses gibberish variable names and a fully deterministic ground truth, ensuring that the task centers purely on structural understanding rather than domain knowledge or memorized patterns. The AI-enabled tool must identify the intended cause-and-effect relationships from simple natural-language descriptions and incorporate them into a partially completed model. To construct each test, the evaluation algorithm begins with a pre-existing causal model made of invented pluralized gibberish nouns. It then generates one or more new causal relationships that should be added to the pre-existing model, again using straightforward, human-sounding templates that avoid complex grammar or misleading phrasing. These synthetic sentences mirror the simplicity and clarity of the causal translation evaluation. The AI-enabled tool is asked to "add the following to my model," and the evaluation checks whether it includes all the required new relationships exactly once, preserves all original relationships, and avoids introducing any additional, unintended ones. This approach provides a controlled and rigorous way to measure an AI-enabled tool's ability to iteratively build causal models. By focusing on simple language, synthetic variables, and deterministic ground truth, the iteration evaluation isolates the model's basic capacity to update an existing causal structure, just as the causal translation evaluation isolates its basic ability to create one from scratch. There are eight (8) tests for qualitative model iteration and nine (9) tests for quantitative model iteration. Tests for model iteration align with the principle to "work with the modeling process, don't bypass it" in keeping with the goal of maintaining interaction in the modeling process.

### 3.1.3 Causal Reasoning

The causal reasoning evaluation (Lynch & Schoenberg, 2026) aims to assess an AI-enabled tools' ability to identify and represent the core causal mechanisms that experts consider essential in real-world systems. Instead of synthetic "alternate universes", this evaluation uses expert-validated sets of variables and their required causal relationships, ensuring that each test focuses on capturing the substance of a domain rather than exact wording or sentence parsing. This is to account for the many possible correct representations of real-world systems. To maintain clarity and avoid unnecessary complexity, the evaluation does not penalize models for adding extra variables or connectors; it simply checks whether all required elements appear with the correct semantics. This design keeps the evaluation focused on whether the model can reproduce the fundamental causal structure that domain experts agree is important. Ground truth for each test is constructed by grouping together variables and causal links that experts identify as central to understanding a system, for example, key factors in pandemic dynamics or organizational change. Each group serves as a "must-have" bundle: the AI-enabled tool receives credit only if it includes every required variable and every required causal relationship from that group. The evaluation procedure is straightforward. The AI-enabled tool is given a short background text describing a real-world situation, and it generates a causal model. The evaluation algorithm then checks whether the model contains the expert-required variables and their corresponding causal connections. If any required part is missing, or if a required relationship is included but with the wrong polarity, the test records a failure for that group. Extra content is ignored, since multiple valid causal models can coexist for the same text. This approach provides a controlled, domain-grounded way to assess whether an AI-enabled tool can capture meaningful causal reasoning rather than simply extract surface patterns. It evaluates the model's ability to express expert-recognized causal mechanisms clearly and correctly, without introducing unnecessary structural challenges or linguistic ambiguity. There are three (3) tests for qualitative causal reasoning and three (3) tests for quantitative causal reasoning. Tests for causal reasoning are aligned with three principles: "do no harm," "develop tools that help people," and "deliver high quality models," through goals of eliminating hallucinated information in generated models and presenting reasoning in terms of feedback and delays.

### 3.1.4 Conformance

The conformance evaluation (Schoenberg et al., 2026) measures an AI-enabled tool's ability to follow explicit modeling instructions when generating a causal model. Rather than checking whether a model matches a single ground truth, this evaluation focuses on whether the AI-enabled tool can respect user-defined constraints, such as including specific variables, producing a minimum number of variables, or generating a certain number of feedback loops. Each test begins with a short background text along with a clear set of structural requirements. The AI-enabled tool is then asked to create a causal model that meets all of these constraints. The evaluation simply checks whether the model satisfies each instruction. If a required variable is missing, if the model is too small or too large, or if it contains too few or too many feedback loops, the test records a failure. Extra variables and extra causal links are allowed unless they violate one of the explicit constraints. This evaluation is intentionally flexible because many different causal structures can satisfy the same instructions. Its purpose is not to enforce one "correct" model but to test whether the AI-enabled tool can reliably shape its output to match what the user asks for, even when the constraints require balancing creativity with structure. By focusing on adherence to instructions, the conformance evaluation provides a simple and direct way to measure an AI-enabled tool's controllability, an essential capability for real modeling workflows where users often need models that follow specific guidelines rather than models that reflect a single fixed structure. There are 18 conformance tests for both qualitative and quantitative engines. Tests for conformance are aligned with the principle to "develop tools that help people" through the goal of adhering to the user's intent.

### 3.1.5 Model Behavior Explanation

The model behavior explanation evaluation assesses an AI-enabled tool's ability to explain why a simulation model behaves the way it does over time. Instead of building or modifying a model, the task focuses on reading an existing model and its Loops that Matter (LTM) feedback-loop dominance analysis (Schoenberg et. al., 2020, Schoenberg et al., 2023), then producing text that correctly reflects the key facts about which feedback loops dominate the model's behavior and when those shifts occur. The evaluation uses a set of known, predetermined facts about each model, such as which loops exist, whether they are reinforcing or balancing, and the time ranges during which each loop drives the system, to create a clear ground truth that the AI-enabled tool must restate. To keep the evaluation simple and controlled, each test provides the AI-enabled tool with both the model and its LTM analysis encoded in JSON (Schoenberg, 2026). The AI-enabled tool is then asked to "explain the behavior of this model," and the evaluation checks whether its explanation contains all the expected factual statements. These facts are concrete and unambiguous (e.g., the number of loops or the specific time at which dominance shifts), avoiding unnecessary linguistic complexity. If any required fact is missing or contradicted, the test records a failure. There are six (6) tests for model behavior explanation that can be applied to discussion engines like Seldon (Schoenberg, 2026). Tests for model behavior explanation are aligned with the principles to "increase access to the modeling process for all" and "develop tools that help people" through goals of helping stakeholders understand model dynamics and presenting reasoning in terms of feedback and delays.

### 3.1.6 Suggested Model Building Steps

The suggested model building steps evaluation assesses an AI-enabled tool's ability to produce appropriate steps for constructing a model when given a problem statement and relevant background knowledge. Instead of generating a model directly, the task focuses on whether the AI-enabled tool can outline the essential building blocks, such as defining stocks, flows, parameters, and relationships, that a modeler would need to construct the model. This test relies on a deterministic set of expert determined ground truth steps. For each scenario, the evaluation provides a short description of the problem and background knowledge, along with a list of canonical steps that should appear in any correct model. These steps are written in plain, straightforward language, avoiding complex phrasing or ambiguous modeling choices. The AI-enabled tool is asked to propose its own set of model-building steps. The evaluation then checks whether each ground-truth step is clearly covered in the generated response using a LLM. The steps do not need to be phrased identically; they must simply capture the same essential modeling idea. If any step is missing, or only partially addressed, the test records a failure. By focusing on step-by-step model construction, this evaluation provides a simple and controlled way to measure whether an AI-enabled tool understands the fundamental structure of standard system dynamics models. There are four (4) tests of suggested model building steps that can be applied to discussion tools. Tests for suggested model building steps align with the principles to "work with the modeling process, don't bypass it" and "complement human abilities, don't substitute for the human" via goals of maintaining interaction in the modeling process and presenting a logical and understandable approach that can be replicated.

### 3.1.7 Suggested Model Fixes

The suggested model fixes evaluation assesses whether an AI-enabled tool can identify and explain formulation errors in models. Instead of repairing the model directly, the task focuses on recognizing mistakes in the model's formulation and the ability of the tool to describe why certain formulations are wrong. Each test provides a model containing known errors along with short background context, and the AI-enabled tool is asked to analyze the model and suggest what errors exist, if any, and how any identified errors should be fixed. The evaluation compares the AI-enabled tool's explanation to a known set of errors, each tied to a specific variable and a clear statement of what is incorrect. The AI-enabled tool does not need to use exact wording, but it must identify the correct variable and accurately convey the nature of the problem. If any expected error is missing, misinterpreted, or not clearly explained, the test records a failure. By focusing on explanation rather than model editing, the evaluation offers a simple and controlled way to measure an AI-enabled tool's ability to diagnose modeling mistakes, an essential capability for supporting modelers in practical modeling workflows. Seven (7) tests for suggested model fixes can be applied to quantitative modeling tools, and seven (7) tests for suggested model fixes can be applied to discussion tools. Tests for suggested model fixes align with the principle to "complement human abilities, don't substitute for the human" by critiquing a model and providing feedback to the modeler.

## 3.2 Alignment with Principles for Evaluation

Table 1 summarizes how the types of tests that have been implemented for each category of evaluation correspond with BEAMS design principles as well as with goals for evaluating AI-enabled tools and their outputs to spur their development in a way that improves dynamic modeling.

Table 1. Principles and goals corresponding with implemented tests

| **Principles** | **Goals** | **Tests** |
| --- | --- | --- |
| Deliver high quality models | Accurately translate causal information; Eliminate hallucinated information in generated models | Causal Translation |
| Work with the modeling process, don't bypass it | Maintain interaction in the modeling process | Model Iteration |
| Do no harm; Develop tools that help people; Deliver high quality models; Use appropriate information in building models | Present reasoning in terms of feedback and delays; Eliminate hallucinated information in generated models | Causal Reasoning |
| Develop tools that help people | Adhere to the user's intent | Conformance |
| Increase access to the modeling process for all; Develop tools that help people | Help stakeholders understand the dynamics of a model; Present reasoning in terms of feedback and delays | Model Behavior Explanation |
| Work with the modeling process, don't bypass it; Complement human abilities, don't substitute for the human | Maintain interaction in the modeling process; Present a logical and understandable approach that can be replicated | Suggested Model Building Steps |
| Complement human abilities, don't substitute for the human | Critique model and provide feedback to modeler | Suggested Model Fixes |

The alignment of these tests with principles demonstrates progress toward all of the principles except to “provide information without bias, stereotypes, or generalizations,” which is an area of ongoing effort in the BEAMS Initiative.

## 3.3 Evaluation of AI-Enabled Tools

Evaluation of AI-enabled tools involves conducting the tests described above for each of the relevant engine types, coupled with different external LLMs. Table 2 summarizes the categories of evaluation for which tests have been implemented, and the number of tests, for each engine type.

Table 2. Summary of tests implemented for each evaluation[3] and engine type

| | **Engine Type** | | |
|---|---|---|---|
| **Evaluation Type** | Qualitative | Quantitative | Discussion |
| Causal Translation | 24 | 9 | |
| Model Iteration | 8 | 9 | |
| Causal Reasoning | 3 | 3 | |
| Conformance | 18 | 18 | |
| Model Behavior Explanation | | | 6 |
| Suggested Model Building Steps | | | 4 |
| Suggested Model Fixes | | 7 | 7 |
| Total Tests | 53 | 46 | 17 |

The performance of AI-enabled tools is summarized in Figures 1-3 and Tables 3-5 below. In Figures 1-3, the average time per test is plotted on the vertical axis and the overall score across tests is plotted on the horizontal axis. In all three figures, optimal performance would be in the lower right of the graph, with a higher score and shorter response time.

Tables 3-5 show the overall performance of each tool along with its performance in each of the categories of evaluation. The overall score is simply the fraction of total tests that a tool passed. Therefore, categories of evaluation with more tests factor more heavily in the overall score. For instance, 24 qualitative causal translation tests comprise nearly half (45%) of the 53 total tests across four evaluation categories for qualitative tools.

[3] https://ub-iad.github.io/sd-ai/#/evals

Figure 1 and Table 3 show the performance of qualitative tools.

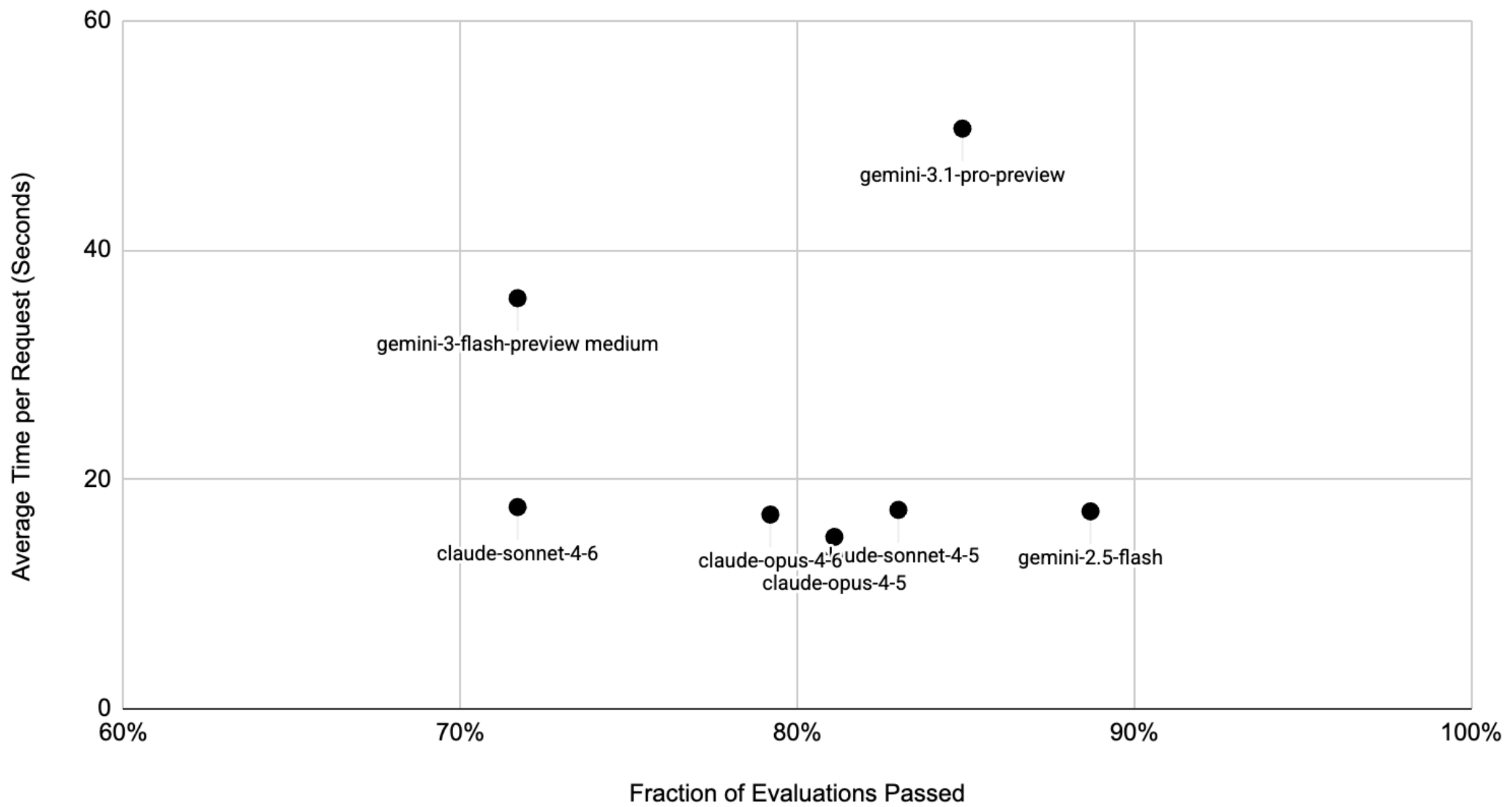

Figure 1. Evaluation of AI-enabled tools for qualitative modeling tasks[4]

Table 3. Performance of AI-enabled tools in categories of qualitative model evaluation

| LLM coupled with Qualitative Engine | Overall Score | Causal Translation | Model Iteration | Causal Reasoning | Conformance |
|---|---|---|---|---|---|
| gemini-2.5-flash | 88.7% | 100% | 87.5% | 33.3% | 83.3% |
| gemini-3.1-pro-preview | 84.9% | 100% | 100% | 0% | 72.2% |
| claude-sonnet-4-5 | 83% | 95.8% | 62.5% | 33.3% | 83.3% |
| claude-opus-4-5 | 81.1% | 87.5% | 75% | 0% | 88.9% |
| claude-opus-4-6 | 79.2% | 95.8% | 62.5% | 33.3% | 72.2% |
| gemini-3-flash-preview medium | 71.7% | 87.5% | 62.5% | 0% | 66.7% |
| claude-sonnet-4-6 | 71.7% | 83.3% | 62.5% | 33.3% | 66.7% |

Figure 2 and Table 4 show the performance of quantitative tools.

[4] https://ub-iad.github.io/sd-ai/#/leaderboard/cld

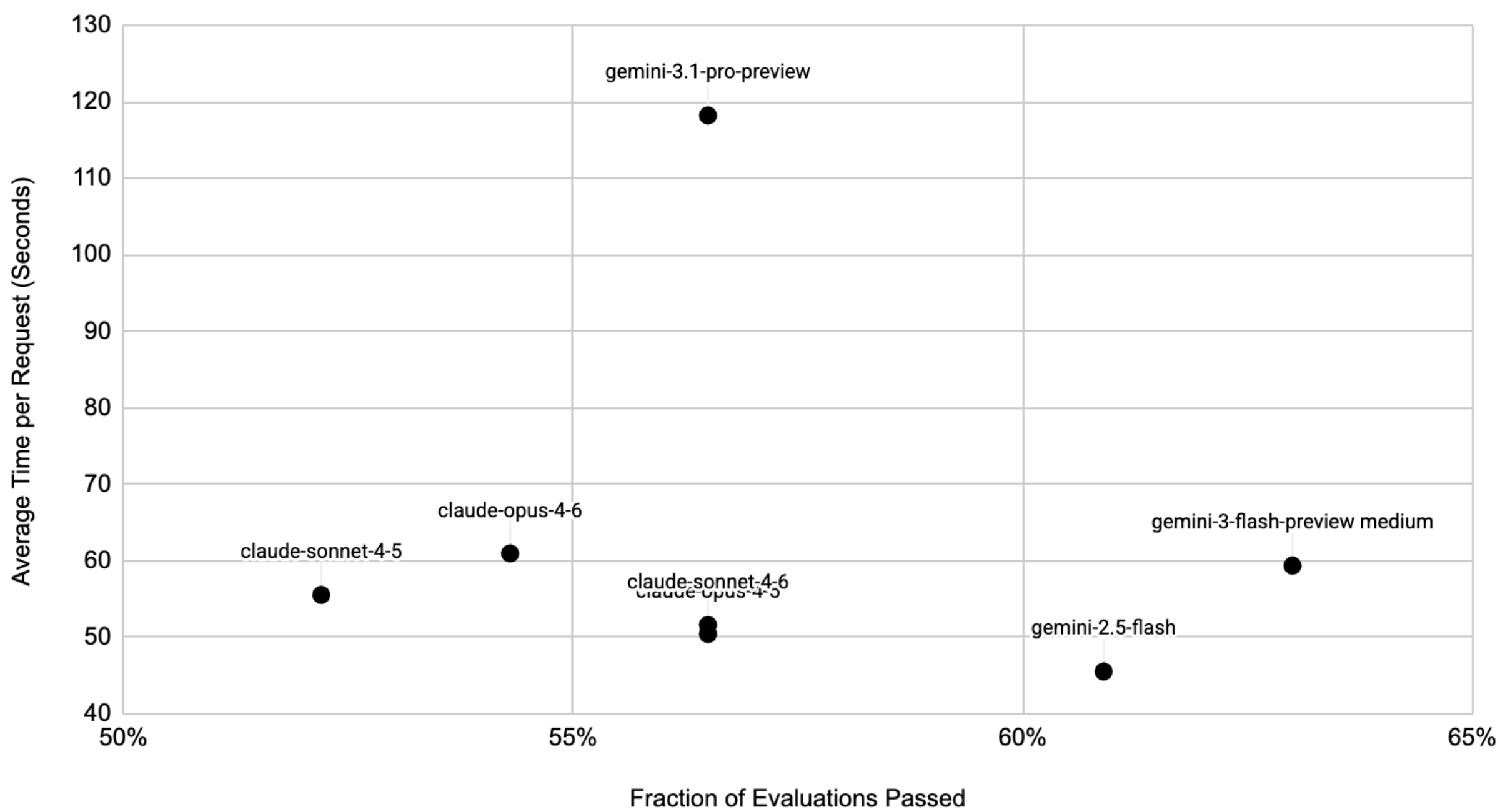

Figure 2. Evaluation of AI-enabled tools for quantitative modeling tasks[5]

Table 4. Performance of AI-enabled tools in categories of quantitative model evaluation

| LLM coupled with Quantitative Engine | Overall Score | Causal Translation | Model Iteration | Causal Reasoning | Conformance | Suggested Model Fixes |
|---|---|---|---|---|---|---|
| gemini-3-flash-preview medium | 63% | 77.8% | 66.7% | 66.7% | 66.7% | 28.6% |
| gemini-2.5-flash | 60.9% | 88.9% | 77.8% | 66.7% | 44.4% | 42.9% |
| gemini-3.1-pro-preview | 56.5% | 77.8% | 88.9% | 66.7% | 50% | 0% |
| claude-opus-4-5 | 56.5% | 88.9% | 77.8% | 33.3% | 44.4% | 28.6% |
| claude-sonnet-4-6 | 56.5% | 88.9% | 44.4% | 33.3% | 66.7% | 14.3% |
| claude-opus-4-6 | 54.3% | 88.9% | 77.8% | 33.3% | 33.3% | 42.9% |
| claude-sonnet-4-5 | 52.2% | 88.9% | 77.8% | 33.3% | 44.4% | 0% |

[5] https://ub-iad.github.io/sd-ai/#/leaderboard/sfd

Figure 3 and Table 5 show the performance of discussion tools.

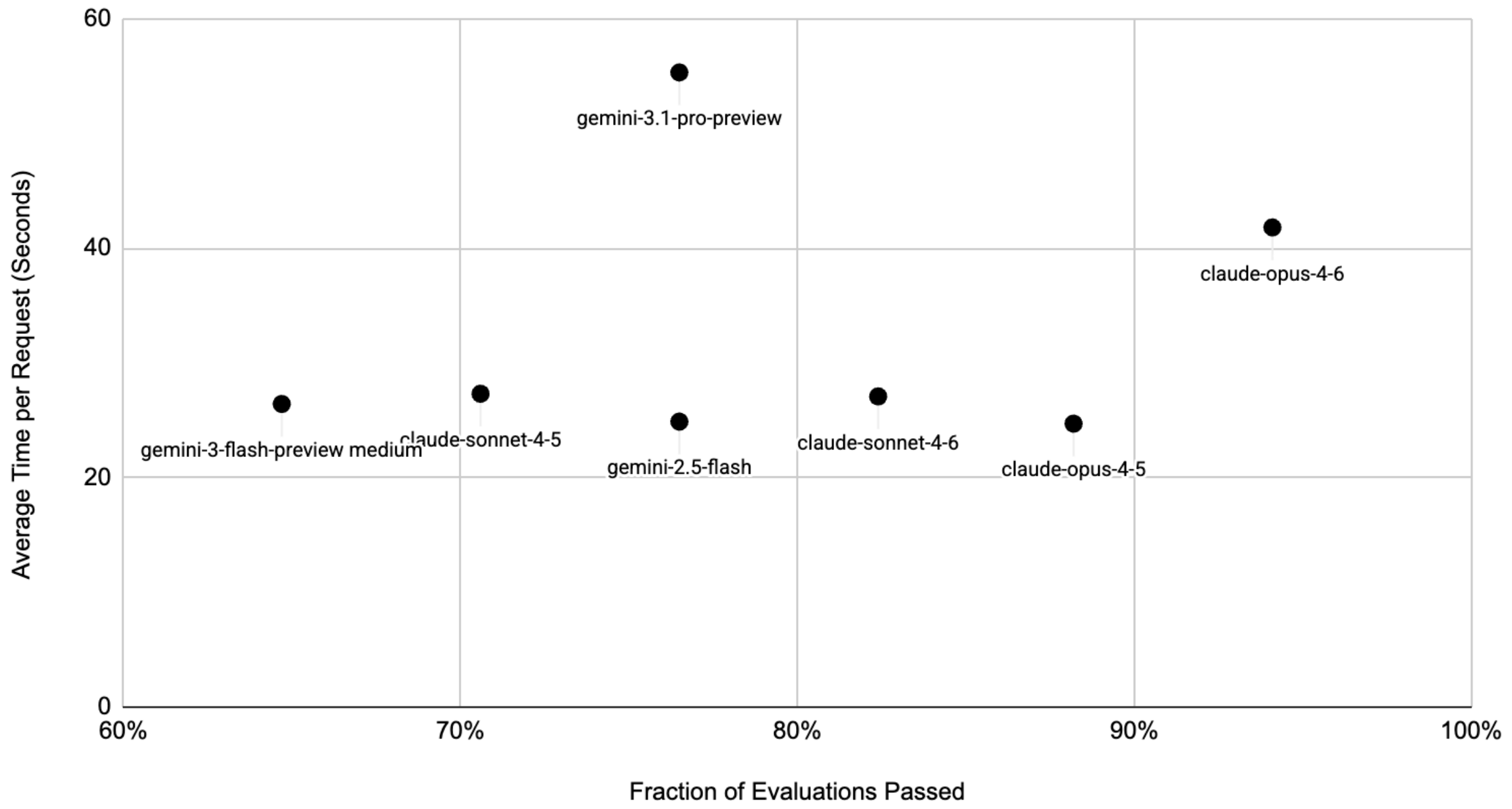

Figure 3. Evaluation of AI-enabled tools for model discussion[6]

Table 5. Performance of AI-enabled tools in categories of evaluation for discussion tasks

| LLM coupled with Seldon Engine | Overall Score | Model Behavior Explanation | Suggested Model Building Steps | Suggested Model Fixes |
|---|---:|---:|---:|---:|
| claude-opus-4-6 | 94.1% | 83.3% | 100% | 100% |
| claude-opus-4-5 | 88.2% | 66.7% | 100% | 100% |
| claude-sonnet-4-6 | 82.4% | 83.3% | 100% | 71.4% |
| gemini-2.5-flash | 76.5% | 66.7% | 75% | 85.7% |
| gemini-3.1-pro-preview | 76.5% | 50% | 75% | 100% |
| claude-sonnet-4-5 | 70.6% | 50% | 100% | 71.4% |
| gemini-3-flash-preview medium | 64.7% | 50% | 50% | 85.7% |

[6] https://ub-iad.github.io/sd-ai/#/leaderboard/discussion

# 4. Discussion

## 4.1 Overall patterns across tasks

Across the BEAMS Initiative's evaluations, AI-enabled modeling tools provided by the sd-ai platform do not perform uniformly; their strengths and weaknesses are highly task-dependent. According to our evaluation data, the clearest, and practically most useful, successes occur when performing discussion tasks. This is consistent with the core capability of large language models (LLMs): transforming information across contexts when the relevant structure is either provided or can be directly inferred from structured inputs (i.e., they are models of language, not of knowledge per se).

On qualitative modeling tasks AI tools generally pass the basic competency tests, causal translation and iteration, and they show the ability to be flexible to the desired level of aggregation of their end users, specifically, they can generally conform to output instructions (e.g., variable inclusion, minimum size, loop counts). These results indicate that, when the task is framed as clear transformations, the LLMs' language modeling strengths are well leveraged. However, within the domain of qualitative modeling tasks, AI tools struggle with causal reasoning tasks grounded in real world examples even though the evaluation scoring tolerates multiple correct formulations. This gap suggests that beyond basic i.e. translation-style mapping, reconstructing expert-recognized mechanisms requires either (a) richer, explicit background information or (b) advancements in the LLMs themselves. Practically, to obtain effective qualitative results, users should include detailed, concrete explanations of the system to be summarized, including salient variables, hypothesized feedback loops, and descriptions of boundary conditions. Doing so re-casts the task as an organization of already-specified content, a task that LLMs excel at.

For quantitative modeling tasks, AI tools show more limited competence. In particular, their performance on the error-fixing task is far worse than Seldon's given the same list of issues to identify and resolve. A practical pathway to improvement is to feed Seldon's output into the quantitative engine as structured evidence. This would create a detect, explain, propose and fix pipeline that narrows the search space and counteracts LLMs' tendency to expand the scope of the model.

Two related findings have important implications for the AI enabled modeling tools we tested. First, no single LLM dominates across all evaluations when it underlies all the tested AI enabled modeling tools. On several evaluations different LLMs can achieve 100%. Second, there is substantial equivalence among the high performing LLMs within each evaluation category. Specifically, several LLMs achieve closely clustered scores. Together, these observations suggest the use of a flexible strategy when it comes to the choice of LLM underlying the tested AI-enabled modeling tools. These results confirm the importance of the design of the sd-ai platform which makes it easy to swap LLMs and they reinforce the necessity of reporting per-evaluation and per-LLM leaderboards rather than a single aggregate score by engine.

Across evaluations, the most modern, "highest-reasoning" and most expensive LLMs were not consistently top performers. A plausible, but untested explanation is that these models "overthink", producing longer, less targeted outputs that increase latency without commensurate accuracy gains. Regardless, for real-world use cases this raises the point that time-to-useful-answer is an important metric. Often an LLM that is "good enough" and fast, rather than the nominally most "capable" model that is slow or verbose is most practical. Observed latencies in the evaluation results reinforce those trade-offs. On average, discussion tasks are fast (~25 seconds per request among high-scoring models), likely because LLMs specialize in reorganizing supplied text, and the output format is not highly structured. Quantitative tasks are slowest (~60 seconds), reflecting longer "thinking" generations and the substantial complexity of structured output the quantitative engine uses. Qualitative tasks are highly variable (from ~17 seconds to 50 seconds among high-scorers), the qualitative engine is likely slower when prompts are longer or require more disambiguation. For interactive modeling sessions, these latency differences materially affect user experience and should inform LLM selection.

The results suggest several concrete design patterns:

1. Front-load described structure for qualitative tasks. Provide explicit variable lists, hypothesized feedback loops, and constraints.
2. Scaffold quantitative model building. Decompose into micro-tasks, small additions, and use tools like Seldon to feed criticisms of generated models back into the LLM for targeted fixes.
3. Leverage discussion engines like Seldon as glue. Use generated model behavior explanation to produce interpretable narratives; use generated modeling steps to drive qualitative/quantitative model construction; and generated suggested fixes as pre-screeners before automated model editing.
4. Optimize for speed under constraints. Prefer LLMs that meet a known accuracy threshold and hit latency targets.
5. Adopt per-task LLM routing. Given the equivalence among top performers and task-specific variability, select the LLMs with the best results for each BEAMS evaluation type.

## 4.2 Areas of Ongoing Effort

The suite of evaluations developed to date through the BEAMS Initiative is an important contribution to the question of how we can start to measure the performance of AI tools for modeling and simulation tasks. The leaderboards for different LLMs coupled with engines available through the sd-ai project could potentially provide an incentive for LLM builders to develop tools that are better suited for the modeling and simulation community. However, the scope of current evaluations is not yet sufficiently comprehensive to ensure that high achievement will truly lead to more responsible and ethical AI tools. While these evaluations align with important principles, additional tests and categories of evaluation will be needed to fully realize the principles for evaluation. And as noted above, tests of the principle to "provide information without bias, stereotypes, or generalizations" have not yet been implemented.

However, ongoing efforts of the BEAMS Initiative point to potential evaluations that would address this gap by considering whether model structures reflecting alternative perspectives are capable of being expressed by an AI-enabled tool.

In line with the overarching mission of developing more responsible and ethical AI tools, collaborators in the BEAMS Initiative have underscored the importance of ensuring that the needs and problems of stakeholders are considered in evaluating AI-enabled tools for modeling and simulation. To work toward this, an ongoing effort is to tie potential evaluations of modeling tasks to distinct use cases that are in turn associated with people who engage with models and/or who are affected by AI-enhanced model output. Moreover, as an open collaboration, the BEAMS Initiative strives to include model stakeholders of all kinds in addition to modelers, software developers, and information scientists.

# 5. Conclusion

The BEAMS Initiative represents an important and timely effort to influence how AI-enabled tools should be developed for modeling and simulation by leveraging knowledge from the modeling and simulation community. By establishing transparent, reproducible, and openly governed benchmarks, the BEAMS Initiative contributes a novel structured framework for assessing whether AI tools can meaningfully support modeling practices, and through this, whether AI tools can help to solve important societal issues. This effort is not only technically significant, but also normatively important: it reinforces that AI systems intended for real-world decision support must be interpretable and aligned with best modeling practices.

By implementing evaluations that span qualitative and quantitative model construction, formulation and design, as well as causal reasoning, model explanation, and error diagnosis, the BEAMS Initiative is working to broaden the scope of what can be measured. These benchmarks move beyond natural language processing assessments toward domain-specific tests that reflect the demands of real modeling work. As a result, evaluations from the BEAMS Initiative expand the field’s ability to rigorously compare tools, understand task-specific strengths and weaknesses, and identify where AI complements human judgement.

Insights from the evaluations implemented through the BEAMS Initiative reveal real and consequential shifts in LLM performance, shifts that would be hard to notice without structured evaluation. The results of this work demonstrate that no single LLM dominates across tasks, that newer and more expensive LLMs do not uniformly outperform smaller and faster models, and that performance depends strongly on task framing. These findings underscore the need for the continued development of task-specific evaluations of AI tools with easily swappable LLMs. As the initiative continues to evolve, BEAMS provides the foundation for a cumulative, community-driven understanding of how AI tools can responsibly and effectively advance the practice of modeling and simulation.